\documentclass{article}
\usepackage{PRIMEarxiv}

\usepackage{amsmath,amssymb,amsfonts}
\usepackage{graphicx}
\usepackage{textcomp}
\usepackage{xcolor}
\def\BibTeX{{\rm B\kern-.05em{\sc i\kern-.025em b}\kern-.08em
    T\kern-.1667em\lower.7ex\hbox{E}\kern-.125emX}}

\usepackage[T1]{fontenc}
\usepackage[utf8]{inputenc}
\usepackage[american]{babel}
\usepackage{csquotes}
\usepackage{microtype}
\usepackage[defblank]{paralist}
\setdefaultenum{(1)}{(a)}{(i)}{A.}
\usepackage[inline]{enumitem}
\usepackage{url}
\usepackage{flushend}
\usepackage{etoolbox}
\usepackage{fix-cm}
\usepackage{textpos}
\usepackage{mfirstuc}
\usepackage{titlecaps}
\usepackage[datesep=.,style=ddmmyyyy]{datetime2}

\usepackage{algorithm}
\usepackage{algpseudocode}
\usepackage{pgf}
\newcommand{\newPerc}[1]{\fpeval{round(#1*100,2)}\%}

\usepackage{tcolorbox}

\newtcolorbox{Summary}{
    sharpish corners, % Better drop shadow
    boxrule = 0pt, % No visible box outline
    toprule = 3.5pt, % Top rule weight
    toptitle = 1mm, % Slight space between top rule and title
    enhanced,
    fuzzy shadow = {0pt}{-2pt}{-0.5pt}{0.5pt}{black!35}, % Shadow settings
    colback = white, % Keep background white
    colframe = gray, % No visible frame
    coltitle=black, % Title color
    fonttitle=\bfseries % Bold font for title
}
\usepackage[framemethod=tikz]{mdframed}

\mdfdefinestyle{mpdframe}{
    frametitlebackgroundcolor   =black!15,
    frametitlerule              =true,
    roundcorner                 =5pt,
    middlelinewidth             =1pt,
    innermargin                 =0.3cm,
    outermargin                 =0.3cm,
    innerleftmargin             =0.3cm,
    innerrightmargin            =0.3cm,
    innertopmargin              =0.5cm,
    innerbottommargin           =0.5cm
}

\tcbuselibrary{skins}
\usetikzlibrary{shadings}

\colorlet{colexam}{red!75!black}

\usepackage{amsmath}
\usepackage{amssymb}
\usepackage{bm}
\usepackage{relsize}
\usepackage{siunitx}
\usepackage[super]{nth}

\usepackage{booktabs}
\usepackage{multirow}
\usepackage{colortbl}
\usepackage{makecell}
\usepackage{rotating}
\usepackage{threeparttable}
\usepackage{adjustbox}

\usepackage{float}
\usepackage{graphicx}
\usepackage{xcolor}
\usepackage{listingsutf8}

\usepackage{xparse}
\usepackage{xspace}

\usepackage[xindy,acronym]{glossaries}
\glsdisablehyper

\usepackage[numbers,sort&compress]{natbib}
\usepackage[hidelinks,bookmarks=false]{hyperref}
\usepackage[capitalise]{cleveref}

\definecolor{mycustomcolor}{HTML}{6D8764}

\tcbset{
  base/.style={
    enhanced,
    colframe=teal!5, % Frame color
    colback=teal!5,  % Background color
    sharp corners,
    borderline west={2pt}{0pt}{teal!70},
    boxrule=0.5mm,
    boxsep=4pt,
    leftrule=2mm,
    titlerule=0pt,
    fonttitle=\bfseries\itshape, % Bold italic title font
    width=\textwidth, % Box width matches text width
    attach boxed title to top left={
      xshift=2mm,
      yshift=-2mm
    }, % Adjust title position
    boxed title style={
      colback=teal!90, % Title background color
      colframe=teal!90, % Title frame color
      sharp corners
    }
  }
}

\newtcolorbox{myexamplec}[2][]{% Two arguments: optional (for labels) and mandatory (for title)
  base,
  title={#2}, % Dynamically set the title
}

\usepackage{subcaption} 
\usepackage{minted}
\usepackage{array}  % already mentioned
\newcolumntype{P}[1]{>{\raggedright\arraybackslash}p{#1}}
\newcommand{\EUProject}{{\tt RoboSAPIENS}\xspace}
\usepackage{hyperref}

\begin{document}

% \title{Digital Twin-based Out of Distribution Detection in Industrial Self-adaptive Robots}\\

%\title{Monitoring and Analysis of Self-adaptive Robots with AI-powered Digital Twins}
% \title{Digital Twins for Dependability Assurance in Self-adaptive Robots}
% \title{Digital Twin-Based Proactive OOD Detection for Self-Adaptive Robotic Systems}
\title{Out of Distribution Detection in Self-adaptive Robots with AI-powered Digital Twins}

% \and
% \and
% \and
% \and
% \and

\author{
\textbf{Erblin Isaku}\textsuperscript{*,\dag}, 
\textbf{Hassan Sartaj}\textsuperscript{*}, 
\textbf{Shaukat Ali}\textsuperscript{*}, \\
\textbf{Beatriz Sanguino}\textsuperscript{\ddag}, 
\textbf{Tongtong Wang}\textsuperscript{\ddag}, 
\textbf{Guoyuan Li}\textsuperscript{\ddag}, 
\textbf{Houxiang Zhang}\textsuperscript{\ddag}, 
\textbf{and Thomas Peyrucain}\textsuperscript{\S} \\
\\
\textsuperscript{*}Simula Research Laboratory, Oslo, Norway \\
\{erblin, hassan, shaukat\}@simula.no \\
\textsuperscript{\dag}University of Oslo, Oslo, Norway \\
\textsuperscript{\ddag}Norwegian University of Science and Technology, Ålesund, Norway \\
\{beatriz.i.s.c.d.c.sanguino, tongtong.wang, guoyuan.li, hozh\}@ntnu.no \\
\textsuperscript{\S}PAL Robotics, Barcelona, Spain \\
thomas.peyrucain@pal-robotics.com
}

% \author{%
%     \IEEEauthorblockN{
%         Erblin Isaku\IEEEauthorrefmark{1}\IEEEauthorrefmark{2},
%         Hassan Sartaj\IEEEauthorrefmark{1},
%         Shaukat Ali\IEEEauthorrefmark{1},
%         Beatriz Sanguino\IEEEauthorrefmark{3},
%         Tongtong Wang\IEEEauthorrefmark{3},
%         Guoyuan Li\IEEEauthorrefmark{3},\\
%         Houxiang Zhang\IEEEauthorrefmark{3},
%         and Thomas Peyrucain\IEEEauthorrefmark{4}
%     }
%     \IEEEauthorblockA{\IEEEauthorrefmark{1}Simula Research Laboratory, Oslo, Norway\\
%     \{erblin, hassan, shaukat\}@simula.no}
%     \IEEEauthorblockA{\IEEEauthorrefmark{2}University of Oslo, Oslo, Norway}
%     \IEEEauthorblockA{\IEEEauthorrefmark{3}Norwegian University of Science and Technology, Ålesund, Norway\\
%     \{beatriz.i.s.c.d.c.sanguino, tongtong.wang, guoyuan.li, hozh\}@ntnu.no}
%     \IEEEauthorblockA{\IEEEauthorrefmark{4}PAL Robotics, Barcelona, Spain\\
%     thomas.peyrucain@pal-robotics.com}
% }

\maketitle
\newcommand{\approach}{{\fontfamily{qhv}\selectfont ODiSAR}}

\begin{abstract}
Self-adaptive robots (SARs) in complex, uncertain environments must proactively detect and address abnormal behaviors, including out-of-distribution (OOD) cases. To this end, digital twins offer a valuable solution for OOD detection. Thus,  
%This is particularly critical in the maritime domain, where deviations from expected dynamics can compromise navigation, mission objectives, or even safety.
we present a digital twin-based approach for OOD detection (\approach{}) in SARs. \approach{} uses a Transformer-based digital twin to forecast SAR states and employs reconstruction error and Monte Carlo dropout for uncertainty quantification. By combining reconstruction error with predictive variance, the digital twin effectively detects OOD behaviors, even in previously unseen conditions. The digital twin also 
%To enable trustworthy self-adaptation, the framework embeds naturally into the MAPLE-K feedback loop, supporting the Monitor and Analyze stages. 
includes an explainability layer that links potential OOD to specific SAR states, offering insights for self-adaptation. We evaluated \approach{} by creating digital twins of two industrial robots: one navigating an office environment, and another performing maritime ship navigation. 
% In both cases, digital twins were used to detect out-of-distribution (OOD) events before they occurred. 
In both cases, \approach{} forecasts SAR behaviors (i.e., robot trajectories and vessel motion) and proactively detects OOD events.
Our results showed that \approach{} achieved high detection performance---up to 98\% AUROC, 96\% TNR@TPR95, and 95\% F1-score---while providing interpretable insights to support self-adaptation. \looseness=-1

\end{abstract}

\keywords{
Self-Adaptive Robots, Digital Twins, OOD Detection, Uncertainty Quantification, Explainable DT}

\section{Introduction}
Autonomous robots are increasingly deployed in dynamic and unpredictable environments, where real-time adaptation to changes in environmental and internal states, as well as mission goals, is critical. This ability, called \textit{self-adaptation}, is key to ensure their dependability and long-term autonomy~\cite{de2013software}. Robots with this capability are known as Self-Adaptive Robots (SARs). \looseness=-1

%In response to this need, the concept of Self-Adaptive Robots (SARs) has emerged as a promising solution to equip autonomous systems with the ability to adjust their behavior in real-time. 

%SARs continuously monitor their internal state and environment and adapt their operation in response to unforeseen or uncertain situations. These adaptive behaviors are commonly realized through architectural frameworks such as MAPE-K (\textbf{M}onitor, \textbf{A}nalyze, \textbf{P}lan, \textbf{E}xecute, \textbf{K}nowledge)~\cite{MAPE-K}. However, recent advances such as the MAPLE-K architecture~\cite{MAPLE-K} proposed in the European ~\EUProject{}~\cite{RoboSAPIENS} project extend MAPE-K by introducing components like \textbf{L}egitimate, which checks the safety and legitimacy of adaptation plans before execution. 

A key challenge in enabling self-adaptation is the timely detection of anomalous conditions. 
Among these, out-of-distribution (OOD), i.e., cases where data deviates from the training distribution~\cite{hendrycks2016baseline, liang2017enhancing}, are critical. While initially studied in image and text classification, OOD is also important in robotics, where data-driven models are often used for perception, planning, and control. OOD cases may arise from scenarios related to sensor faults, actuator drift, or environmental changes on which the robot wasn't trained~\cite {amodei2016concrete, cai2020real}. Thus, early OOD detection is essential for dependable robot behavior in dynamic environments. 

Although well-studied in machine learning and vision tasks~\cite{yang2024generalized, ren2019likelihood, hendrycks2016baseline, ma2024research}, OOD still has gaps in the SAR context. 
\begin{inparaenum}
    \item Most OOD approaches are reactive, detecting anomalies only after they occur. In SARs, proactive detection, i.e., forecasting future SAR states and identifying OOD before they occur, is critical for timely planning and safe adaptation~\cite{ji2022proactive}.
    \item OOD approaches often give binary outputs without explaining why a sample is anomalous, limiting trust and hindering integration with self-adaptation frameworks, e.g., MAPLE-K \cite{MAPLE-K}, which may require reasoning about the cause before triggering adaptation.
    \item Most approaches rely on forecasting error, reconstruction error, or uncertainty, whereas combining these can yield robust OOD detection, as emphasized in~\cite{hendrycks2016baseline, ma2024research}.
    \item Though increasingly used in robotics, Digital Twins (DTs) often serve as passive simulators. Few works explore their role in adaptive decision-making~\cite{azari2025self, mo2025digital}, while almost no work is on interpretable and proactive OOD detection using DTs.
    % \item Finally, most evaluations target single-domain systems, limiting understanding of cross-domain generalizability. 
\end{inparaenum}

To address these gaps, we present \approach{}, a DT-based OOD detection approach integrated with the MAPLE-K self-adaptive architecture \cite{MAPLE-K}. \approach{} combines sequence-to-sequence forecasting, reconstruction-based error, and predictive uncertainty via Monte Carlo Dropout for proactive OOD detection of anomalous future states. It also supports state-level attribution, identifying which SAR states are most associated with anomalous behavior for interpretability.

We evaluate \approach{} on two industrial SAR case studies from the European \EUProject{} project \cite{RoboSAPIENS}. 
The first, from NTNU's industrial partner---Kongsberg Maritime AS\footnote{\url{https://www.kongsberg.com/maritime/}}---a Norwegian company, involves autonomous vessels navigating at sea and using motion prediction to support onboard decision-making for safe and efficient operation.
The second case from PAL Robotics\footnote{\url{https://pal-robotics.com/}} involves indoor robots autonomously navigating dynamic environments subject to blocked paths, changing layouts, and unexpected human presence. 
% is about autonomous maritime vessels adapting predictive models for motion estimation under varying environmental uncertainties, such as wind, waves, and currents. 
In both, \approach{} forecasts system behaviors (i.e., robot trajectories and vessel motion states) and proactively detects OOD cases.

%For the vessels case, we simulate environmental disturbances, such as wind, waves, and currents, across three maneuver types: \textit{Zigzag}, \textit{Random}, and \textit{Turning}, each with varying intensity levels. In the PAL Robotics case, we introduce sensor-level disturbances by injecting noise into the robot’s odometry data during waypoint-based navigation tasks. These disturbances create controlled out-of-distribution conditions to test the predictive accuracy and assess the robustness of our approach under different operational contexts. 

Our results show that \approach{} consistently outperforms a baseline (i.e., RMSE) that relies on forecasting error as the only indicator for OOD detection, achieving up to 98\% AUROC, 96\% TNR@TPR95, and 95\% OOD F1-score for maritime vessel, and 96\%, 94\%, and 89\%, respectively, for mobile robot---while also providing confidence-aware predictions and interpretable attributions at the SAR state level. 
In most maritime vessel scenarios, the majority of predictions fall into either the IND Confident (59–62\%) or OOD Uncertain (33–39\%) categories. In contrast, the mobile robot case study shows an almost equal split between IND Confident (48.75\%) and OOD Confident (51.25\%), indicating potential overconfidence. These differences highlight an opportunity for future work on balancing model confidence through uncertainty regularization or calibration techniques.

\section{Application Context and Industrial Case Studies}
\begin{figure}[htbp]
\vspace{0.3em}
  \centering
    \includegraphics[width=0.7\columnwidth]{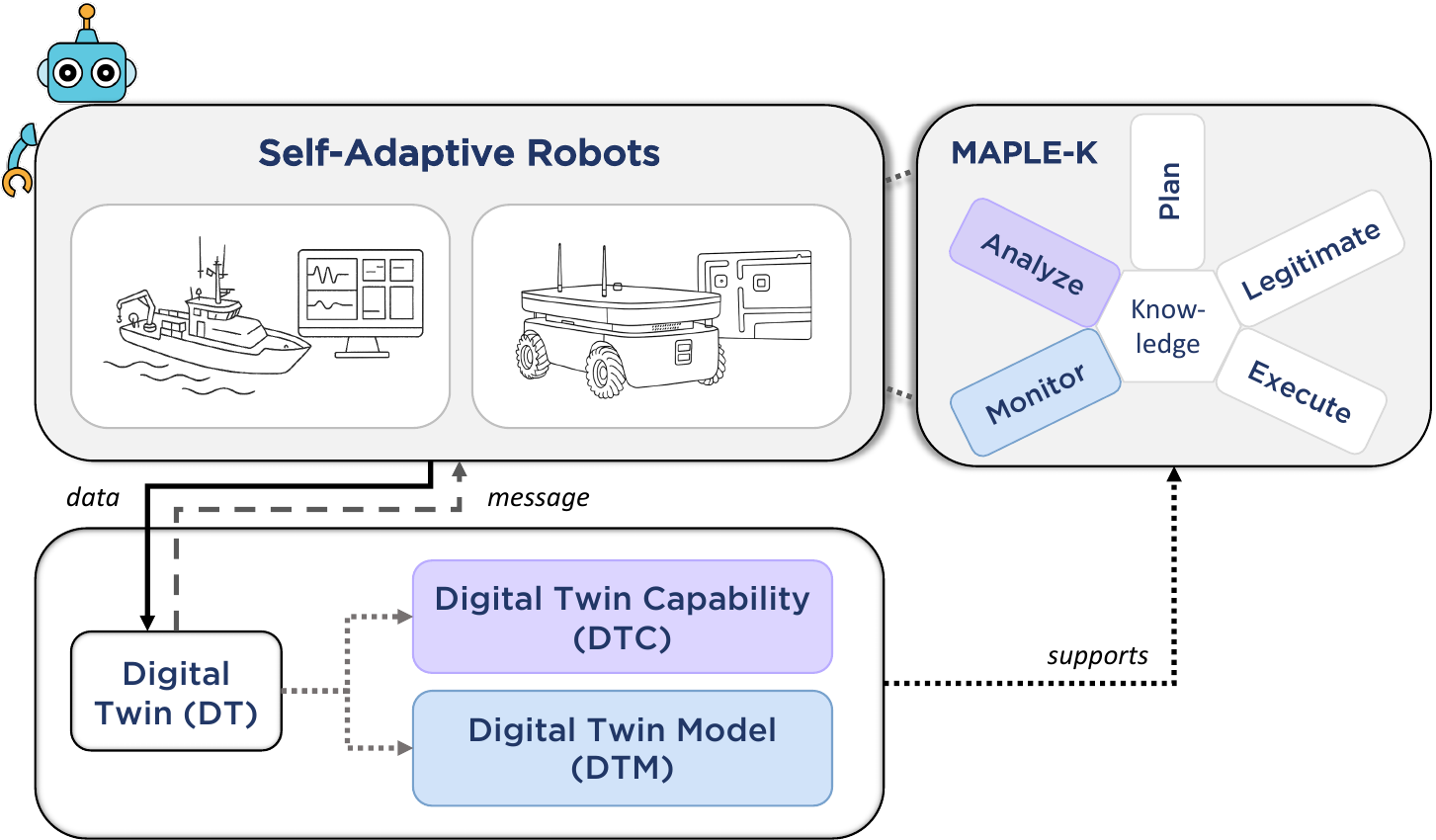}
    \caption{\approach{} in the context of the MAPLE-K loop---implemented inside self-adaptive robots. The DT, composed of DTM and DTC components, supports the Monitor and Analyze phases (shown in light colors), respectively, enabling interpretable OOD detection and adaptation within the MAPLE-K framework.}
    \label{fig:maple-k}
\end{figure}

This work is part of a European project, ~\EUProject{}~\cite{RoboSAPIENS}, whose objective is to build autonomous SARs that can operate in unknown environments and adapt their behavior in response to unforeseen situations. To this end, \EUProject{} proposes MAPLE-K architecture~\cite{MAPLE-K} for implementing self-adaptations in robots, also shown in Figure~\ref{fig:maple-k}, which is an extension of the classical MAPE-K architecture~\cite{MAPE-K}. In MAPLE-K, the \textbf{M}onitor component continuously monitors the state of the robot as well as its environment and analyzes it using methods implemented in the \textbf{A}nalyze component. If the analysis results reveal that adaptation is necessary (e.g., an anomaly is detected), the \textbf{P}lan component generates a set of adaptation plans to be executed and selects the most suitable one. The \textbf{L}egitimate component, newly introduced in MAPLE-K, verifies whether the adaptation is safe to execute. Once this verification is passed, the adaptation is executed by the \textbf{E}xecute component. The \textbf{K}nowledge component collects relevant information from all the MAPLE-K components.
The dynamic nature of SARs, driven by various changes such as structural (e.g., adding or upgrading sensors), functional (e.g., updating navigation algorithms), and environmental (e.g., diverse weather and human-robot interactions), introduces significant complexity in implementing self-adaptation and can cause the adaptation space to expand drastically~\cite{MAPLE-K}. 
This presents a critical challenge for industry practitioners to effectively monitor and analyze the behavior of SAR under continuously changing conditions. 
To address this, we propose a digital twin-based approach that enables real-time detection and adaptive response to OOD cases under varying unpredictable operational contexts.

A digital twin, in our case, consists of two components (see Figure \ref{fig:maple-k}): a digital twin model (DTM), which is a replica of a SAR, and a digital twin capability (DTC), which is an additional feature provided by the digital twin (e.g., OOD) that uses the DTM along with the most up-to-date state of the robot to perform the capability. Once the DTC detects unexpected behavior (e.g., a potential OOD likely to occur in the near future, in our context), it sends a signal to the SAR to initiate adaptation planning using the \textit{Plan} component in MAPLE-K. Essentially, the digital twin in our setup supports two components of MAPLE-K implemented in SARs: continuous monitoring of the SARs’ states with the \textit{Monitor} component and analyzing them to signal any potential OOD with the \textit{Analyze} component.

In the context of this paper, we evaluate \approach{} using two industrial SAR case studies from \EUProject{}. 
The first case study focuses on autonomous vessels (AVs) provided by Kongsberg Maritime that is a global leader in building maritime applications. In this case, we explore how AVs' digital twins can enable them to detect OOD while navigating at sea.
The second case study is provided by PAL Robotics (Spain), a world leader in building service robots. Here, we demonstrate the application of \approach{} with a robot operating in an office environment. Its digital twin helps OOD detection.

%...
%\approach{} operates in alignment with the MAPLE-K loop, where the digital twin model supports the Monitor phase by forecasting future vessel states, and the digital twin capability supports the Analyze phase by identifying and interpreting deviations from expected behavior.

\section{Approach}
\begin{figure*}[tbp]
% \vspace{0.3em}
  \centering
    \includegraphics[width=1\textwidth]{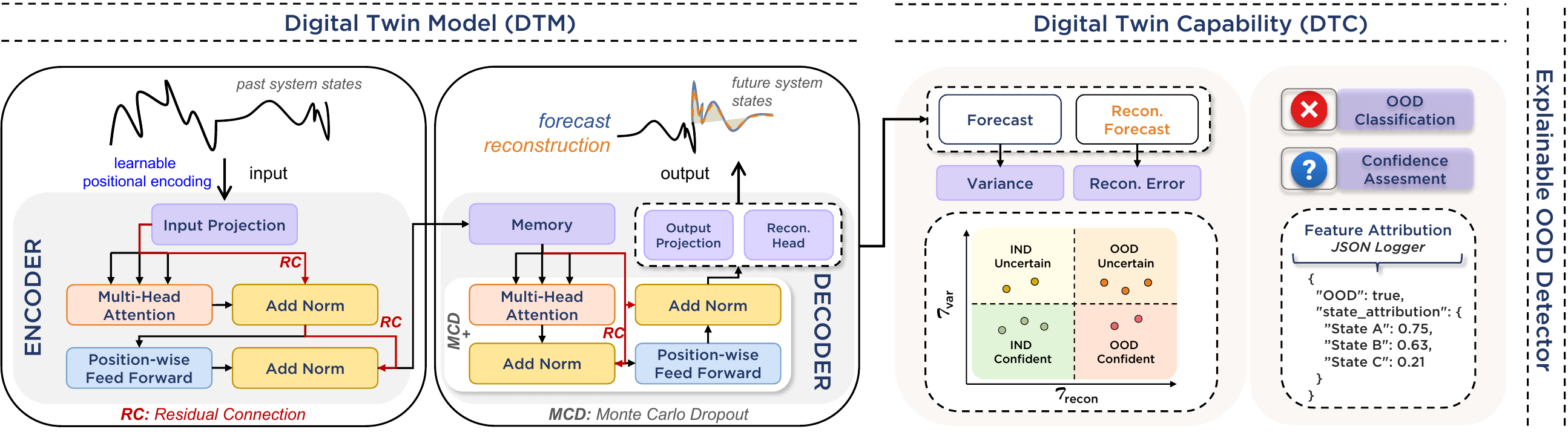}
    \caption{Overview of the proposed Digital Twin-based approach (\approach{}) for proactive OOD detection.
    The Transformer-based Digital Twin Model (DTM) takes historical input and produces both forecasted and reconstructed future system states. The Digital Twin Capability (DTC) analyzes these outputs using reconstruction error and uncertainty (via MC Dropout) to detect potential OOD instances. The Explainable OOD Detector combines this analysis to flag OOD states and attributes them to the most contributing system features.}
    \label{fig:approach-overview}
\end{figure*}

\Cref{fig:approach-overview} presents a high-level overview of \approach{}, our proactive and interpretable OOD detection approach based on the principles of the MAPLE-K loop for enabling self-adaptation. 
The approach is composed of two key components:
\begin{inparaenum}[(i)]
    \item the digital twin model (DTM) responsible for forecasting the expected robot behavior, and
    \item the digital twin capability (DTC) responsible for detecting and interpreting deviations from the normal operational behavior. 
\end{inparaenum} 
The full implementation of the proposed approach is publicly available on GitHub.\footnote{\url{https://github.com/Simula-COMPLEX/ODiSAR}} 

% \Cref{fig:approach-overview} illustrates the step-by-step data-driven pipeline for proactive and explainable out-of-distribution detection. 
\approach{} takes a sequence of past system states as input (e.g., sensor readings and control commands) and creates the DTM consisting of an encoder and decoder to forecast future system states while simultaneously reconstructing them. The encoder processes the input using self-attention mechanisms and positional encoding to capture temporal dependencies. This encoded representation is passed to a decoder, which generates both forecasted outputs and their reconstructions through two parallel heads. To estimate predictive uncertainty, Monte Carlo (MC) Dropout is applied at inference, producing multiple stochastic forward passes. The resulting forecasts are post-processed by the DTC, which computes the reconstruction error and the forecast variance. These two scores are compared against learned thresholds to classify each prediction as either in-distribution or out-of-distribution, and as confident or uncertain. Finally, for predictions flagged as OOD, the approach provides a state-level attribution, identifying the most contributing states and exporting this information in a JSON format. This resulting information enables domain experts to continuously monitor and analyze abnormal behaviors of SARs and take proactive measures to ensure their dependability.

% \begin{figure*}[tbp]
% % \vspace{0.3em}
%   \centering
%     \includegraphics[width=1\textwidth]{figures/ase-final-overview-diagram.pdf}
%     \caption{Overview of the proposed Digital Twin-based approach (\approach{}) for proactive OOD detection.
%     The Transformer-based Digital Twin Model (DTM) takes historical input and produces both forecasted and reconstructed future system states. The Digital Twin Capability (DTC) analyzes these outputs using reconstruction error and uncertainty (via MC Dropout) to detect potential OOD instances. The Explainable OOD Detector combines this analysis to flag OOD states and attributes them to the most contributing system features.}
%     \label{fig:approach-overview}
% \end{figure*}

\subsection{SAR Data and Model Settings}

To determine the most suitable architecture for the DTM, we conducted an exploratory pilot study comparing several time series forecasting models, including recurrent neural networks (RNNs), variational autoencoders (VAEs), generative adversarial networks (GANs), and transformer neural networks.  Based on preliminary training stability and forecasting results, the transformer model consistently outperformed the other models. Therefore, we adopted a vanilla transformer as the base model for the DTM and extended it with a reconstruction head to support proactive OOD detection.

The input and output features are selected based on domain-specific knowledge and task, specifically motion and trajectory prediction. For the autonomous maritime vessel case study, we forecast ship dynamics, including surge/sway velocity, yaw rate, and roll angle/rate. For the autonomous mobile robot case, we forecast the robot's future position states.
To optimize both model architecture and training performance, we used the Optuna framework~\cite{akiba2019optuna} for automated hyperparameter tuning. Table~\ref{tab:dtm_params} presents the final configurations derived from the results of the pilot experiment for each case study.

\begin{table}[h]
\centering
\caption{\approach{}: Model and Training Parameters}
\label{tab:dtm_params}
\begin{tabular}{lcc}
\toprule
\textbf{Parameter} & \textbf{Maritime Vessel} & \textbf{Mobile Robot} \\
\midrule
Model Dimension ($d_{\text{model}}$) & 64 & 64 \\
Number of Attention Heads & 4 & 4 \\
Feedforward Dimension & 128 & 128 \\
Dropout Rate & 0.1 & 0.2 \\
Batch Size & 16 & 16 \\
Learning Rate & 0.0001 & 0.0001 \\
Epochs & 200 & 200 \\
\bottomrule
\end{tabular}
\end{table}

\subsection{Creating DTM}

The purpose of DTM is to learn the normal behavior of the target system through time series forecasting, thereby supporting the \textit{Monitor} phase in the MAPLE-K control model by generating future predictions of key system variables (e.g., yaw rate, velocity, or position) and enabling continuous observation of expected behavior.
Therefore, we formulate this forecasting task as a sequence-to-sequence prediction problem. Given a sequence of past observations 
\( \mathbf{X}_{t-w+1:t} = [\mathbf{x}_{t-w+1}, \dots, \mathbf{x}_{t}] \) where $t$ denotes the current time step, the goal is to predict the corresponding future sequence of system states 
\( \hat{\mathbf{Y}}_{t+1:t+h} = [\hat{\mathbf{y}}_{t+1}, \dots, \hat{\mathbf{y}}_{t+h}] \), 
where \( w \) is the input window size and \( h \) is the forecast horizon.

We design the DTM using a Transformer-based architecture composed of an encoder and decoder with multi-head attention and feedforward layers. 
The model learns how the system behaves/maneuvers over time by analyzing patterns in past data and uses this understanding to predict future states.
In addition to forecasting, the model is trained to reconstruct its own forecasted outputs. 
% This reconstruction task is handled by a lightweight multi-layer perceptron (MLP) head attached to the decoder output.
This is achieved via a lightweight multi-layer perceptron (MLP) reconstruction head attached to the decoder output. 
By reconstructing what the model has just predicted, it learns to form a robust internal representation of what "normal" predictions look like. 
% This representation is later used by the DTC to detect abnormal patterns without needing access to future ground truth during deployment.
This representation is later used by the DTC to proactively detect abnormal patterns during deployment, based on how well the model reconstructs its own state predictions.

% To train the reconstruction component
% As part of training, the model minimizes a reconstruction loss between the forecasted outputs \( \hat{\mathbf{Y}}_{t+1:t+h} \) and their reconstructed versions \( \tilde{\mathbf{Y}}_{t+1:t+h} = [\tilde{\mathbf{y}}_{t+1}, \dots, \tilde{\mathbf{y}}_{t+h}] \). The reconstruction loss is calculated using~\Cref{eq:recon-loss}:

As part of training, the model minimizes two objectives. The first is a \emph{forecasting loss} that compares the predicted future sequence \( \hat{\mathbf{Y}}_{t+1:t+h} \) to the ground truth future values \( \mathbf{Y}_{t+1:t+h} \). This loss is typically computed using Mean Squared Error (MSE):

\begin{equation}
\mathcal{L}_{\text{Forecast}} = \frac{1}{h} \sum_{i=1}^{h} \left\| \hat{\mathbf{y}}_{t+i} - \mathbf{y}_{t+i} \right\|^2
\label{eq:forecast-loss}
\end{equation}

The second is a \emph{reconstruction loss} between the forecasted outputs and their reconstructed versions \( \tilde{\mathbf{Y}}_{t+1:t+h} = [\tilde{\mathbf{y}}_{t+1}, \dots, \tilde{\mathbf{y}}_{t+h}] \). The reconstruction loss, similar to forecasting, is computed as:

\begin{equation}
\mathcal{L}_{\text{Recon}} = \frac{1}{h} \sum_{i=1}^{h} \left\| \tilde{\mathbf{y}}_{t+i} - \hat{\mathbf{y}}_{t+i} \right\|^2
\label{eq:recon-loss}
\end{equation}
% The overall training objective combines both the forecasting and reconstruction tasks. Specifically, the total loss is calculated using~\Cref{eq:total-loss}. 

The overall training objective combines both the forecasting and reconstruction losses. Specifically, the total loss is calculated using:

\begin{equation}
\mathcal{L}_{\text{Total}} = \underbrace{\mathcal{L}_{\text{Forecast}}}_{\text{Forecasting Loss}} + \underbrace{\mathcal{L}_{\text{Recon}}}_{\text{Reconstruction Loss}}
\label{eq:total-loss}
\end{equation}

This training setup enables the DTM to learn to forecast future system states and reconstruct those jointly. The dual objective ensures that the model develops a robust internal representation of normal system behavior, which is needed for supporting proactive OOD detection at deployment time.
% \begin{equation}
% \mathcal{L}_{\text{Total}} = \underbrace{\text{MSE}(\hat{\mathbf{Y}}, \mathbf{Y})}_{\text{Forecasting Loss}} + \underbrace{\mathcal{L}_{\text{Recon}}}_{\text{Reconstruction Loss}}
% \label{eq:total-loss}
% \end{equation}

% Here, \( \mathbf{Y} = [\mathbf{y}_{t+1}, \dots, \mathbf{y}_{t+h}] \) denotes the ground truth future sequence, used as the target for forecasting.

% The input and output features are selected based on domain knowledge and vary across use cases. For the autonomous vessel dataset, we forecast ship dynamics such as surge/sway velocity, yaw rate, and roll angle/rate. For the autonomous mobile robot dataset, we forecast the robot’s future position states.

% To select optimal hyperparameters for each case study, we leveraged the Optuna framework for automated hyperparameter tuning~\cite{akiba2019optuna}. Table~\ref{tab:dtm_params} summarizes the model and training configurations used for both case studies.

% \begin{figure}[ht]
%     \centering
%     \begin{subfigure}{\columnwidth}
%         \includegraphics[width=\linewidth]{figures/ntnu-forecasted.png}
%         \caption{NTNU: Ship Motion Forecasting}
%     \end{subfigure}
%     \hfill
%     \begin{subfigure}{\columnwidth}
%         \includegraphics[width=\linewidth]{figures/pal-forecasted.png}
%         \caption{PAL: Robot Trajectory Forecasting}
%     \end{subfigure}
%     \caption{DTM forecasting examples for (a) NTNU and (b) PAL Robotics use cases.}
%     \label{fig:dtm_forecast_examples}
% \end{figure}

\subsection{Creating DTC}

The DTC enables proactive OOD detection by analyzing the forecasted outputs of the DTM. It includes two key metrics---reconstruction error (\cref{eq:recon-loss}) and predictive uncertainty (\cref{eq:variance}), i.e., variance---to identify deviations from expected behavior. 
% This design eliminates the need for future ground truth during deployment, enabling real-time detection of abnormal system behavior.
At inference time, the DTC evaluates how well the model can reconstruct its own predictions. A high reconstruction error indicates that the forecasted sequence lies outside the distribution the model has learned, suggesting potential OOD behavior.
In addition, we estimate predictive uncertainty using MC Dropout~\cite{gal2016dropout}. During inference, we perform multiple stochastic forward passes through the DTM with dropout enabled. Each pass produces a slightly different forecast due to the randomness introduced by dropout. By computing the variance across these sampled forecasts using~\Cref{eq:variance}, we capture the model’s epistemic uncertainty. In~\Cref{eq:variance}, \( \hat{\mathbf{y}}^{(n)}_{t+i} \) denotes the forecasted output at time step \( t+i \) from the \( n^\text{th} \) stochastic forward pass, and \( \bar{\mathbf{y}}_{t+i} \) is the mean of these forecasts over \( N \) passes.

\begin{equation}
\text{Var}(\hat{\mathbf{y}}_{t+i}) = \frac{1}{N} \sum_{n=1}^{N} \left( \hat{\mathbf{y}}^{(n)}_{t+i} - \bar{\mathbf{y}}_{t+i} \right)^2
\label{eq:variance}
\end{equation}

To support proactive detection, the DTC defines two thresholds in~\Cref{eq:thresholds} based on statistics from in-distribution validation data. Here, \( \mu \) and \( \sigma \) are the mean and standard deviation of the reconstruction errors and variances, and \( k \) is a tunable sensitivity parameter (e.g., \( k = 3 \) for a 3-sigma threshold~\cite{pukelsheim1994three}).

\begin{equation}
    \tau_{\text{recon}} = \mu_{\text{recon}} + k \cdot \sigma_{\text{recon}}, \quad
    \tau_{\text{var}} = \mu_{\text{var}} + k \cdot \sigma_{\text{var}}
\label{eq:thresholds}
\end{equation}

Each forecast window—i.e., the entire sequence of predicted future values over the forecast horizon—is then assigned to one of four categories (illustrated inside DTC block in Figure~\ref{fig:approach-overview}) based on these classification categories:
\begin{inparaenum}[(i)]
    \item \textit{IND and Confident:} Low reconstruction error, low uncertainty
    \item \textit{IND and Uncertain:} Low reconstruction error, high uncertainty
    \item \textit{OOD and Uncertain:} High reconstruction error, high uncertainty
    \item \textit{OOD and Confident:} High reconstruction error, low uncertainty.
\end{inparaenum}

This categorization supports interpretable, confidence-aware, and actionable detection of anomalous behavior by not only flagging potential OOD events but also indicating the \approach{}'s level of certainty. For instance, uncertain cases can undergo further monitoring or human review, while confident OOD detections can trigger immediate adaptive responses. Similarly, uncertain IND cases might cause warrant caution or continued observation.

\subsection{Explainable OOD Detection}

To complement detection with interpretability, we enhance the DTC with an explainable mechanism that identifies which system features contributed most to the OOD classification. This provides actionable insight during deployment and supports human operators in understanding anomalous behavior.

For each forecast window flagged as OOD (i.e., when the reconstruction error exceeds the threshold), we compute the feature-wise root mean squared error (RMSE) between the model's forecast \( \hat{\mathbf{y}} \) and its reconstruction \( \tilde{\mathbf{y}} \) using~\Cref{eq:rmse}. Here, \( h \) is the forecast horizon length, and \( \mathbf{e}_{\text{feat}} \in \mathbb{R}^{d} \) is a vector of reconstruction errors for each of the \( d \) output features.

\begin{equation}
e^{(i)}_{\text{feat}} = \sqrt{ \frac{1}{h} \sum_{j=1}^{h} \left( \tilde{y}^{(i)}_{j} - \hat{y}^{(i)}_{j} \right)^2 }, \quad \text{for } i = 1, \dots, d
\label{eq:rmse}
\end{equation}

We then rank the features (i.e., states) by their RMSE values and report the top three as an interpretable attribution of which system states showed the strongest anomalous responses. These attributions are saved in a structured JSON file alongside metadata such as the sequence index, start/end time steps, reconstruction error, uncertainty variance, and assigned OOD category.

Each forecast window is categorized into one of four semantic quadrants (e.g., IND Confident or OOD Uncertain) based on its reconstruction error and uncertainty level. For instance, in the example shown in Figure~\ref{fig:json_example}, the window is labeled as \texttt{"red"}, corresponding to the \textit{OOD \& Confident} region in the quadrant plot. 
The \texttt{state\_attribution} field highlights the top three states---\textit{Surge Speed}, \textit{Sway Speed}, and \textit{Yaw Rate}---that show the highest reconstruction errors and were most impacted by the underlying anomaly.

\begin{figure}[ht]
\centering
\begin{minted}[fontsize=\small, linenos=false]{json}
{
  "sequence_index": 3,
  "start_time_step": 420,
  "end_time_step": 479,
  "is_OOD": true,
  "reconstruction_error": 0.17066404223442078,
  "uncertainty_variance": 0.018417222425341606,
  "recon_exceeds_threshold": true,
  "uncertainty_exceeds_threshold": false,
  "category": "red",
  "state_attribution": {
    "Surge Speed": 0.26233699917793274,
    "Sway Speed": 0.21531985700130463,
    "Yaw Rate": 0.13875150680541992
  }
}
\end{minted}
\caption{Example of structured JSON output for a forecast window (autonomous maritime vessel case study) flagged as OOD. The top-3 contributing states are provided under \texttt{state\_attribution}.}
\label{fig:json_example}
\end{figure}

While our approach does not follow formal explainability frameworks (e.g., SHAP~\cite{lundberg2017unified} or LIME~\cite{ribeiro2016should}), it aims to improve practical interpretability by highlighting the most affected system states showing potential anomalous behavior. This lightweight diagnostic capability is designed to assist human operators and practitioners in tracing the root of behavioral deviations in SARs.

\section{Experiment Design}

\begin{table*}[ht]
\centering
\caption{
Overview of the experimental setup across both cases studies. 
Maneuver types marked with $^\dagger$ indicate cases where disturbances were added to generate OOD data for inference and model evaluation.
}
\renewcommand{\arraystretch}{1.9}
\setlength\extrarowheight{2pt}
\begin{tabular}{>{\centering\arraybackslash}m{2.3cm} >{\centering\arraybackslash}m{2.4cm} >{\arraybackslash}m{3.8cm} >{\arraybackslash}m{2.2cm} >{\arraybackslash}m{2.5cm}}
\hline
\textbf{Use Case} & \textbf{Task} & \textbf{Disturbance/OOD Source} & \textbf{Maneuver Type} & \textbf{Variants} \\
\hline
\multirow{3}{*}{\raisebox{-.4\totalheight}{\includegraphics[width=2.7cm]{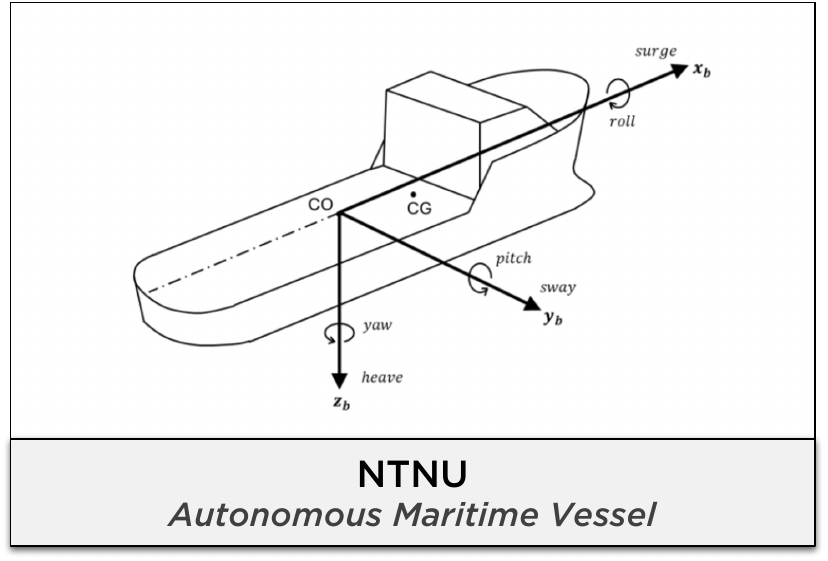}}} &
\multirow{3}{*}{\raisebox{-.4\totalheight}{\includegraphics[width=2.7cm]{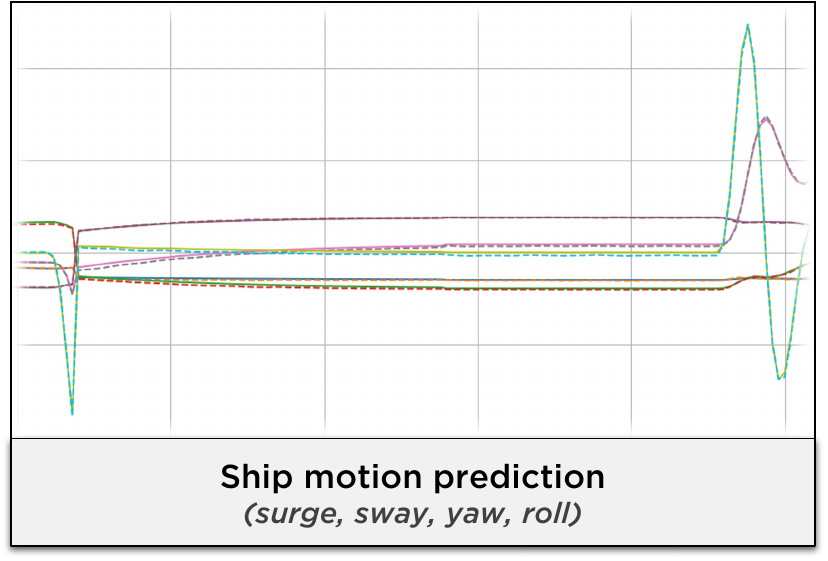}}} &
\multirow{3}{*}{\makecell[{{l}}]{Environmental disturbance\\ (wind, waves, currents)}} &
Zigzag$^{\dagger}$ & 10°, 15°, 20°, 30° \\
& & & Random$^{\dagger}$ & 1, low, high \\
& & & Turning$^{\dagger}$ & 10°, 15°, 20°, 30° \\
\hline
\multirow{3}{*}{\raisebox{-.4\totalheight}{\includegraphics[width=2.7cm]{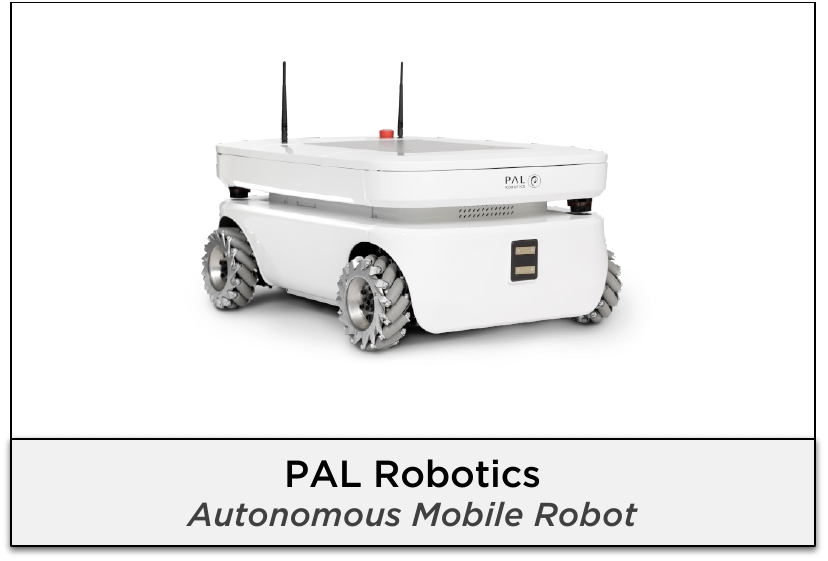}}} &
\multirow{3}{*}{\raisebox{-.4\totalheight}{\includegraphics[width=2.7cm]{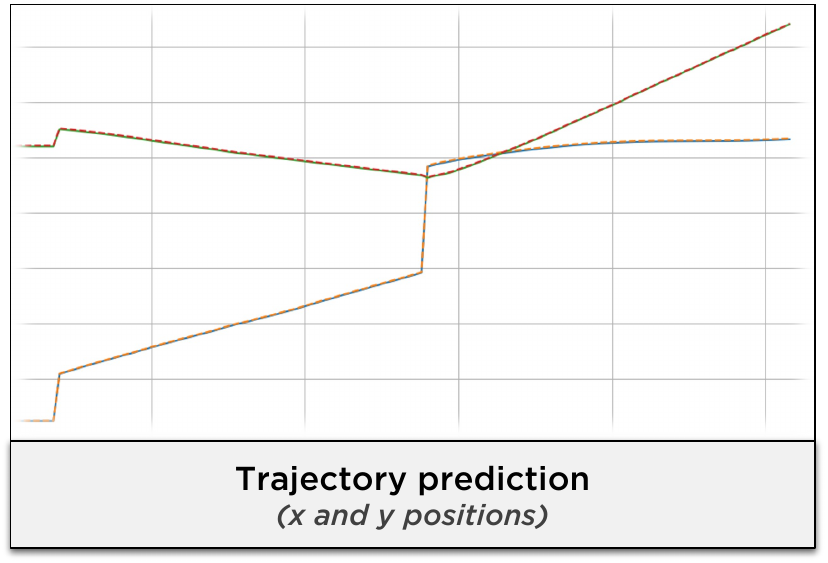}}} &
\multirow{3}{*}{\makecell[{{l}}]{Sensor noise\\ ($/odom$ topic)}} &
\multirow{3}{*}{Waypoint$^{\dagger}$} & \multirow{3}{*}{--} \\
& & & & \\
& & & & \\
\hline
\end{tabular}
\label{tab:usecase-summary}
\end{table*}

\subsection{Simulation Environment \& Robotic System}
\paragraph*{Autonomous Maritime Vessel}
The data used in this study was collected using a professional navigation bridge
simulator, called K-Sim Navigation, manufactured by Kongsberg Maritime AS~\cite{KSimNavigation}. This
simulator is highly accurate and realistic, allows for the simulation of a wide range of environmental conditions, and is widely used for training nautical students and professional captains.
For our experiments, we used a Ro-Ro ferry (Ferry Basto Fosen) model equipped with two azimuth thrusters -- one at the bow and one at the stern. During simulations, thrusters operated at a constant speed of 206 RPM with a fixed propeller pitch of 80\%, and only the thruster angles were adjusted to perform maneuvers.

\paragraph*{Autonomous Mobile Robot}  
We used the open-source PAL Robotics OMNI Base Simulation environment~\cite{pal_robotics_simulator}, which integrates with ROS~2 and Gazebo to provide a realistic 3D simulation of the TIAGo OMNI Base robot~\cite{tiago_omni_base}. 
This robot features omnidirectional 3-DOF planar motion (x, y, $\theta$) and is equipped with two LIDAR sensors that provide a full 360º field of view for real-time obstacle detection and autonomous indoor navigation tasks.
It is widely used in research and industry applications such as office automation, healthcare, hospitality, and logistics, due to its precise navigation and versatile sensing capabilities. 
In our study, the simulation was conducted within PAL’s office map environment, where the robot performed waypoint-based navigation.

\subsection{Maneuvers and Disturbance Scenarios}
\paragraph*{Autonomous Maritime Vessel}

A variety of maneuvers were conducted and recorded over a 20-minute period for each maneuver, with data sampled at a frequency of 1 Hz. As shown in Table~\ref{tab:usecase-summary}, these maneuvers included Zigzag, Turning, and Random patterns, each with several variants. To simulate OOD conditions, environmental disturbances such as wind, waves, and currents were introduced. The recorded data captures key navigational variables, including surge and sway velocities, yaw rate, roll dynamics, and environmental factors like wind, waves, and currents.

To evaluate the \approach{}'s ability to detect different types of OOD events, three environmental disturbance cases were designed.
In each case, disturbances were introduced gradually between minutes 7 and 14 to simulate realistic transitions, designated as the OOD event. 
In Case 1 and Case 3, the initial 7 minutes had no disturbances. Starting from minute 7, wind (e.g., 18 knots from 45° with gusts and directional variation), waves, and current (e.g., 6 knots from 205°) were applied, with smooth transitions in direction (90°/min) and speed (35–40 knots/min). After minute 14, only wind gusts (Case 1) or no disturbances (Case 3) remained. In Case 2, the simulation began with light wind and waves, intensified mid-way to stronger wind and current, and returned to initial conditions after minute 14.

\paragraph*{Autonomous Mobile Robot}  

The simulated robot followed a series of waypoint-based navigation tasks within the PAL Robotics office environment using the ROS~2 Nav2 stack. For each task, five spatially separated waypoints were sampled from the map, ensuring minimum clearance from obstacles and coverage across the environment. The robot navigated sequentially to each waypoint while maintaining 3-DOF planar motion.

To simulate sensor degradation and evaluate how well \approach{} performs under such situations, noise was added to the robot's odometry after reaching the first waypoint. Specifically, Gaussian noise with zero mean was applied to the position, orientation, and velocity values in the $/mobile\_base\_controller/odom$ topic. This noisy period lasted for 80 seconds and is treated as the OOD event. During the entire navigation task, odometry data were recorded at 10~Hz.

% \subsection{Datasets and Preprocessing}

\subsection{Evaluation Metrics}
To evaluate the performance of \approach{}, we use three commonly adopted metrics in OOD detection~\cite{che2021deep, hsu2020generalized}: the Area Under the Receiver Operating Characteristic (AUROC), the True Negative Rate at 95\% True Positive Rate (TNR@TPR95), and the F1-score. These metrics allow us to assess both the general separability of in-distribution (IND) vs. OOD samples and the model's behavior at specific decision thresholds. 
\\

\paragraph*{AUROC}  
AUROC evaluates the model's ability to distinguish IND from OOD samples across all thresholds, measuring the trade-off between true positive rate (TPR) and false positive rate (FPR):
\begin{align}
  \text{AUROC} &= \int_0^1 \text{TPR}(\alpha) \, d(\text{FPR}(\alpha)),
\end{align}
where \(\alpha\) is the threshold. A value closer to 1.0 indicates better separability.

\paragraph*{TNR@TPR95}  
This metric evaluates how well the model reduces false positives while maintaining a TPR of 95\%. The threshold \(\alpha^*\) is computed as:
\begin{align}
  \alpha^* &= \arg\min_{\alpha} \left| \text{TPR}(\alpha) - 0.95 \right|, \\
  \text{TNR@TPR95} &= 1 - \text{FPR}(\alpha^*).
\end{align}

\paragraph*{F1-score}  
The F1-score balances precision and recall, and is defined as:
\begin{equation}
  \text{F1} = 2 \cdot \frac{\text{Precision} \cdot \text{Recall}}{\text{Precision} + \text{Recall}}
\end{equation}
where:
\[
\text{Precision} = \frac{\text{TP}}{\text{TP} + \text{FP}}, \qquad \text{Recall} = \frac{\text{TP}}{\text{TP} + \text{FN}}
\]

A high F1-score indicates that the model performs well in detecting OOD instances while minimizing false positives and false negatives.

\subsection{Research Questions}
\label{sec:study:rqs}
Our experiment investigates the following research questions (RQs).

% \begin{itemize}
%     \item [\textbf{RQ1}] \emph{How effective is \approach{} in detecting OOD behavior caused by environmental disturbances such as wind, waves, and ocean currents?}
%     % Notes:
%     % 1- all cases
%     % 2- cases individually (case 1, case 2, and case 3)
%     % 3- by maneuver type
%     \item [\textbf{RQ2}] \emph{To what extent can \approach{} generalize across different types of vessel maneuvers?}
%     % Notes:
%     % to answer this we train the model on individual maneuver types and test on the rest
%     \item [\textbf{RQ3}] \emph{How effectively does \approach{} support the interpretation of OOD events through confidence-aware categorization and feature attribution?}
%     % Notes:
%     % 1- Quantify how often orange/red zones align with true OOD events.
%     % 2- Analyze whether attributed features match real root causes (e.g., Surge Speed spiking during malfunctions).
%     \item [\textbf{RQ4}] \emph{How does the reconstruction-based thresholding in \approach{} compare to other OOD detection methods like Mahalanobis distance or RMSE?}
% \end{itemize}

\begin{itemize}

    % \item[\textbf{RQ2}] \emph{To what extent can \approach{} generalize across different maneuver types?}

    % \item[\textbf{RQ3}] \emph{How does \approach{} compare to a baseline forecasting model using RMSE-based thresholding for standard OOD detection?}

    % \item[\textbf{RQ3}] \emph{How does \approach{} compare to an OOD detection method based solely on forecasting error?}
    \item[\textbf{RQ1}] \emph{How effective is \approach{} in detecting out-of-distribution behavior under varying disturbance sources?}
    \item[\textbf{RQ2}] \emph{How confident is \approach{} in its predictions under different operational conditions?}    
    \item[\textbf{RQ3}] \emph{How does \approach{} compare with a forecasting-error-only OOD detection method?}
\end{itemize}

% To evaluate the effectiveness of our proposed method, we focus on two core research questions.
RQ1 assesses whether \approach{} can reliably detect out-of-distribution (OOD) behavior under varying disturbance sources, both environmental (e.g., wind, waves, and currents) and sensor-based (e.g., odometry topic). While the autonomous mobile robot case lacks data diversity, we conduct a small-scale robustness study in the autonomous maritime vessel case by training on one representative scenario (i.e., random maneuvering) and evaluating across others.

RQ2 explores how confidently \approach{} operates under different conditions. Although the model doesn’t produce explicit confidence scores, we estimate predictive uncertainty using Monte Carlo Dropout by measuring the variance of forecasted outputs. This enables us to assess the reliability of predictions and support more risk-aware OOD detection.

RQ3 compares \approach{} with a baseline that uses only forecasting error (RMSE) for OOD detection. This helps determine whether integrating forecasting, reconstruction, and uncertainty estimation leads to measurable gains over simpler, threshold-based alternative.

\section{Results and Analyses}
\paragraph{RQ1 - Effectiveness}

\begin{table*}[ht]
\centering
\caption{OOD detection performance across training setups and evaluation scenarios. The results illustrate the effectiveness of \approach{} (RQ1) and its comparison with an RMSE-based baseline (RQ3), using AUROC, TNR@TPR95, and F1-scores for IND and OOD samples.}
\label{tab:rq1-results-merged}
\renewcommand{\arraystretch}{1.1}
\resizebox{\textwidth}{!}{%
\begin{tabular}{lllcccccccc}
\toprule
\multirow{2}{*}{\textbf{Trained On}} 
& \multicolumn{2}{c}{\textbf{Evaluated On}} 
& \multicolumn{4}{c}{\textbf{\approach{}}} 
& \multicolumn{4}{c}{\textbf{RMSE Baseline}} \\
\cmidrule(lr){2-3} \cmidrule(lr){4-7} \cmidrule(lr){8-11}
& Category & Scenario 
& AUROC & TNR@TPR95 & \multicolumn{2}{c}{F1-Score} 
& AUROC & TNR@TPR95 & \multicolumn{2}{c}{F1-Score} \\
\cmidrule(lr){6-7} \cmidrule(lr){10-11}
& & & & & IND & OOD & & & IND & OOD \\

\midrule
\multirow{7}{*}{All Maneuvers}
  & \multirow{3}{*}{\textit{OOD Case}} 
    & Case 1 & \newPerc{0.9788} & \newPerc{0.9555} & \newPerc{0.96} & \newPerc{0.93}  
             & \newPerc{0.9442} & \newPerc{0.6419} & \newPerc{0.43} & \newPerc{0.59} \\
  & & Case 2 & \newPerc{0.9862} & \newPerc{0.9673} & \newPerc{0.96} & \newPerc{0.95}  
             & \newPerc{0.9700} & \newPerc{0.9220} & \newPerc{0.50} & \newPerc{0.66} \\
  & & Case 3 & \newPerc{0.9742} & \newPerc{0.9495} & \newPerc{0.97} & \newPerc{0.94}  
             & \newPerc{0.9451} & \newPerc{0.9334} & \newPerc{0.46} & \newPerc{0.62} \\
  \cmidrule(lr){2-3}
  & \multirow{3}{*}{\textit{Maneuver}} 
    & Zigzag & \newPerc{0.9845} & \newPerc{0.9638} & \newPerc{0.97} & \newPerc{0.95} 
             & \newPerc{0.9655} & \newPerc{0.9491} & \newPerc{0.88} & \newPerc{0.85} \\
  & & Random & \newPerc{0.9783} & \newPerc{0.9615} & \newPerc{0.97} & \newPerc{0.94} 
             & \newPerc{0.9562} & \newPerc{0.9140} & \newPerc{0.57} & \newPerc{0.65} \\
  & & Turning & \newPerc{0.9755} & \newPerc{0.9511} & \newPerc{0.97} & \newPerc{0.95} 
              & \newPerc{0.9660} & \newPerc{0.9381} & \newPerc{0.10} & \newPerc{0.55} \\
  \cmidrule(lr){2-3}
  & \textit{Overall} 
    & All data combined & \newPerc{0.9753} & \newPerc{0.9510} & \newPerc{0.96} & \newPerc{0.94} 
                        & \newPerc{0.9543} & \newPerc{0.9269} & \newPerc{0.54} & \newPerc{0.64} \\
\midrule
\multirow{7}{*}{Random}
  & \multirow{3}{*}{\textit{OOD Case}} 
    & Case 1 & \newPerc{0.9857} & \newPerc{0.9555} & \newPerc{0.94} & \newPerc{0.90} 
             & \newPerc{0.9537} & \newPerc{0.8599} & \newPerc{0.30} & \newPerc{0.56} \\
  & & Case 2 & \newPerc{0.9833} & \newPerc{0.9368} & \newPerc{0.96} & \newPerc{0.94} 
             & \newPerc{0.9710} & \newPerc{0.9220} & \newPerc{0.11} & \newPerc{0.58} \\
  & & Case 3 & \newPerc{0.9745} & \newPerc{0.9190} & \newPerc{0.95} & \newPerc{0.92} 
             & \newPerc{0.9382} & \newPerc{0.9334} & \newPerc{0.16} & \newPerc{0.55} \\
  \cmidrule(lr){2-3}
  & \multirow{3}{*}{\textit{Maneuver}} 
    & Zigzag & \newPerc{0.9825} & \newPerc{0.9491} & \newPerc{0.97} & \newPerc{0.95} 
             & \newPerc{0.9415} & \newPerc{0.9358} & \newPerc{0.29} & \newPerc{0.60} \\
  & & Random & \newPerc{0.9873} & \newPerc{0.9271} & \newPerc{0.92} & \newPerc{0.89} 
             & \newPerc{0.9591} & \newPerc{0.9140} & \newPerc{0.00} & \newPerc{0.53} \\
  & & Turning & \newPerc{0.9802} & \newPerc{0.9394} & \newPerc{0.95} & \newPerc{0.93} 
              & \newPerc{0.9682} & \newPerc{0.9381} & \newPerc{0.12} & \newPerc{0.55} \\
  \cmidrule(lr){2-3}
  & \textit{Overall} 
    & All data combined & \newPerc{0.9796} & \newPerc{0.9286} & \newPerc{0.94} & \newPerc{0.90} 
                        & \newPerc{0.9423} & \newPerc{0.9254} & \newPerc{0.22} & \newPerc{0.56} \\
\midrule
Waypoint & \textit{Maneuver} & Waypoint & \newPerc{0.9652} & \newPerc{0.9440} & \newPerc{0.91} & \newPerc{0.89} 
                        & \newPerc{0.9656} & \newPerc{0.8804} & \newPerc{0.84} & \newPerc{0.84} \\

\bottomrule
\end{tabular}
}
\end{table*}

% \Cref{tab:rq1-results-merged} presents the OOD detection performance of \approach{} under varying disturbance sources, evaluated across three training setups:
% \begin{inparaenum}[i)]
%     \item all maneuver types from the NTNU use case,
%     \item only random maneuver from NTNU, and
%     \item a waypoint-based navigation scenario from the PAL Robotics use case.
% \end{inparaenum}
\Cref{tab:rq1-results-merged} presents the OOD detection performance of \approach{} across two case studies. For the Autonomous Maritime Vessel, the evaluation includes various maneuver types such as Zigzag, Turning, and Random, each tested under multiple disturbance scenarios. For the Autonomous Mobile Robot, performance is assessed in a Waypoint-based indoor navigation setting under sensor noise perturbations.

When trained on all maneuvers, \approach{} achieves strong and consistent results across both OOD cases and maneuver types. AUROC scores remain above 97\%, OOD F1-scores range from 93\% to 95\%, and TNR@TPR95 stays above 94\%. These results highlight the model’s effectiveness across diverse and unseen disturbance patterns. The overall combined performance (AUROC 97.53\%, OOD F1-score 94\%, TNR@TPR95 95.1\%) further confirms its effectiveness under varied test conditions.

To assess how well \approach{} generalizes from a limited but diverse training set, we also consider a setup where the model is trained solely on random maneuvers, which include a mix of dynamic behaviors without following a fixed pattern.
Even under this constraint, the model maintains high detection performance, with AUROC values consistently above 97\% and OOD F1-scores between 89\% and 94\%, and TNR@TPR95 ranging from 91.9\% to 94.9\%. This demonstrates that \approach{} can generalize well to previously unseen maneuver types, despite a narrower training distribution.

For the waypoint navigation scenario (PAL Robotics), training and evaluation are both limited to a single maneuver type. Here, \approach{} achieves an AUROC of 96.52\%, TNR@TPR95 of 94.4\%, and an OOD F1-score of 89\%. While generalization cannot be assessed in this setting, the results confirm reliable performance within the scenario.

Overall, the results indicate that \approach{} is highly effective in detecting OOD behavior under various disturbances. Generalization is strongest when training data covers a diverse set of maneuvers, but the approach still shows strong performance even under more constrained training setups.

\begin{tcolorbox}[colframe=black!50, colback=gray!5, boxrule=0.3mm]
\approach{} achieves high OOD detection performance across varied disturbance sources, with AUROC scores above 97\%, TNR@TPR95 up to 96\%, and OOD F1-scores up to 95\% when trained on diverse maneuvers. It maintains robust performance even under limited training diversity. In the PAL Robotics use case, results remain solid, though generalizability is less conclusive.
\end{tcolorbox}

% \input{rq1n3-table-results}
% \input{rq1-table-results}
% \input{rq2-table-results}

% Notes:

\paragraph{RQ2 - Confidence}
\begin{figure*}[tbp]
  \centering
  \includegraphics[width=1\textwidth]{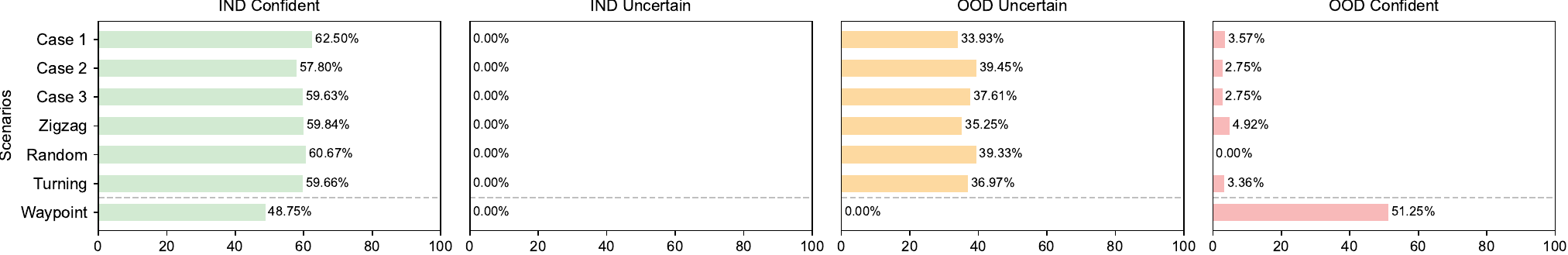}
  \caption{Distribution of model confidence across different operational scenarios. Each bar reflects the proportion of predictions falling into one of four categories---IND Confident, IND Uncertain, OOD Uncertain, and OOD Confident---based on reconstruction error and predictive variance. The $Waypoint$ scenario is from the autonomous mobile robot case study, while the remaining scenarios originate from the autonomous maritime vessel case. This highlights how the model's confidence varies with scenario type and detection outcome.}
  \label{fig:rq2-result}
\end{figure*}

To address RQ2, we examine how confidently the model behaves under varying operational conditions. Using predictive variance estimated via MC Dropout, we categorize model outputs into four quadrants based on whether reconstruction error and forecast variance exceed their respective thresholds: IND Confident, IND Uncertain, OOD Uncertain, and OOD Confident. This categorization provides deeper insights into the reliability and consistency of the model’s predictions across scenarios.

\Cref{fig:rq2-result} presents the distribution of these confidence categories across several scenarios. The autonomous maritime vessel case study includes Cases 1–3 (as OOD sources), Zigzag, Random, and Turning (as maneuver types), while the Waypoint scenario originates from the autonomous mobile robot case.

Across the vessel scenarios, most predictions fall into the IND Confident category ($\geq 57\%$), aligning with expectations, as the model is trained on these in-distribution data. In contrast, predictions labeled as OOD are predominantly uncertain, with OOD Uncertain rates ranging between $33$-$39\%$, while OOD Confident predictions remain consistently low ($<5\%$). This pattern highlights a key insight: uncertainty is effective at flagging a lack of confidence but insufficient on its own for reliable OOD detection. In most OOD cases, the model correctly flags anomalies but expresses low certainty -- showing that reconstruction error remains the more decisive indicator. Therefore, uncertainty is best used as a supportive metric, not a standalone criterion.

The Waypoint scenario (autonomous mobile robot) shows a contrasting trend -- $51\%$ of predictions are OOD Confident, and only $48.75\%$ are IND Confident, with no OOD Uncertain or IND Uncertain cases. While this suggests high certainty across predictions, results from RQ1 reveal lower OOD detection performance for mobile robot data, implying that the DT model may be overconfident. This calls for a deeper investigation into the model’s calibration and generalization under limited training diversity and unseen operational settings, highlighting a potential direction for future work, such as uncertainty regularization or calibration methods.

\begin{tcolorbox}[colframe=black!50, colback=gray!5, boxrule=0.3mm]
\approach{} shows high confidence in in-distribution data, as expected given its training exposure. Uncertainty helps flag unreliable predictions but is insufficient alone for OOD detection, reinforcing the need for reconstruction error as the main indicator. 
% In the PAL Robotics scenario, the absence of uncertainty suggests overconfidence, likely due to limited training diversity.
\end{tcolorbox}

\paragraph{RQ3 - Baseline Comparison}
We compare \approach{} against a baseline that relies solely on forecasting error (RMSE) for OOD detection, across multiple scenarios and training configurations (see Table~\ref{tab:rq1-results-merged}).

At a high level, \approach{} consistently outperforms the baseline across all evaluation dimensions. While the RMSE-based method shows relatively high AUROC scores (often >90\%), its threshold-based metrics (TNR@TPR95 and F1-Score) are significantly lower and highly inconsistent -- particularly for OOD detection. For example, under the “All Maneuvers” training setup, the RMSE baseline achieves an AUROC of 95.43\% but an OOD F1-score of only 64\%. This contrast highlights a key limitation of relying on forecasting error alone for OOD detection.

The discrepancy arises from how each metric operates.
% \begin{inparaenum}[a)]
%     \item AUROC evaluates performance across all possible thresholds, measuring overall separability between IND and OOD distributions. It reflects potential, but not practical performance.
%     \item TNR@TPR95 and F1-score, on the other hand, depend on a fixed threshold. These metrics reflect real-world effectiveness but are sensitive to threshold selection and class imbalance.
% \end{inparaenum}
While AUROC reflects the \approach{}’s ability to discriminate between IND and OOD instances across all possible thresholds, metrics like F1-score and TNR@TPR95 evaluate performance at a fixed threshold. As such, AUROC represents the best case scenario assuming an optimal cutoff, whereas F1-score and TNR@TPR95 indicate how well the model performs with the actual decision boundary in use, being more representative of practical performance.

We notice that the RMSE baseline is highly sensitive to the choice of threshold, and its performance drops significantly in scenarios where RMSE values of IND and OOD samples are not well separated, or when there is class imbalance.
For instance, in the “Random” training setup, OOD F1-score drops to 56\% overall and even lower in specific cases.

In contrast, \approach{} achieves consistently high performance across all metrics, with OOD F1-scores above 89\% and TNR@TPR95 above 91\% in all setups. This highlights its robustness, particularly in handling varied disturbances and training conditions.

Overall, while the RMSE-based baseline may appear promising in terms of AUROC, its poor calibration and reliance on error magnitudes make it unreliable without extensive threshold tuning. \approach{}, by combining reconstruction and uncertainty thresholds, delivers both strong separability and practical detection reliability across all test conditions.

\begin{tcolorbox}[colframe=black!50, colback=gray!5, boxrule=0.3mm]
\approach{} consistently offers high performance across metrics and scenarios, in contrast to the forecasting-only baseline, which shows unstable results due to its sensitivity to threshold selection and class imbalance. This demonstrates the value of combining reconstruction error with uncertainty for robust and reliable OOD detection.
\end{tcolorbox}

\section{Threats to Validity}

This section discusses potential threats that may affect the validity of our results and findings.

\textbf{Internal Validity.} 
A potential internal validity threat can be due to the selection of hyperparameters. Small variations in these parameters---such as learning rate or dropout rate---can significantly affect forecasting accuracy and OOD detection outcomes.  To mitigate this, we conducted a pilot study using automated hyperparameter optimization (i.e., Optuna), allowing us to systematically explore parameter space. Final configurations were selected based on consisten performance across validation sets. 
% Although we employed automated hyperparameter tuning (e.g., using Optuna), the model's performance might still be sensitive to specific parameter settings. Variations in these settings could lead to different OOD detection performance, potentially influencing our findings. To mitigate this, we conducted extensive parameter optimization and reported configurations for reproducibility.

\textbf{External Validity.} 
The generalizability of our findings is subject to constraints associated with our evaluation datasets. The datasets used were obtained from specific simulation scenarios involving maritime vessels and mobile robot systems. Although these datasets realistically represent self-adaptive robot systems, the observed results may vary when the method is applied to other domains or operational environments. Future research should investigate the applicability of our approach across broader contexts to strengthen external validity.

\textbf{Construct Validity.} 
Construct validity concerns the extent to which our metrics accurately measure proactive out-of-distribution detection capability and interpretability. We addressed this by selecting widely recognized and established metrics such as AUROC, F1-score, and TNR\@TPR95. However, alternative metrics or additional qualitative assessments might offer complementary insights into system performance and interpretability.

\textbf{Conclusion Validity.} 
Our conclusions are supported by consistent results across two diverse robotic systems and a range of OOD scenarios. We applied the same methodology and architecture with minimal tailoring to each case, and we report multiple performance metrics to capture different aspects of model effectiveness. 
Nevertheless, repeated evaluations, additional datasets, and anomaly scenarios could further validate and enhance the reliability of our results.

\section{Discussion and Lessons Learned}
Our evaluation across two diverse robotic systems---an autonomous maritime vessel (NTNU) and an autonomous mobile robot (PAL Robotics)--- demonstrates that \approach{} offers an effective and robust approach for proactive OOD detection.
Based on our analysis, we present the following key lessons learned. 

\paragraph{Relevance to self-adaptive control loops}
\approach{} supports key components of self-adaptive robotic architectures, such as MAPE-K~\cite{MAPE-K}, AWARE~\cite{sanwouo2025breaking}, and MAPLE-K~\cite{MAPLE-K} in our context. Specifically, it enhances the Monitor phase through continuous forecasting and state reconstruction, and Analyze phase via proactive and interpretable OOD detection. Furthermore, by providing confidence-aware predictions, the proposed approach contributes to building actionable Knowledge, enabling informed adaptation decisions. 
For example, in the autonomous maritime vessel case study, we observed that roll-related states (i.e., roll angle and roll rate) were among the most frequently identified contributors to OOD classifications. This aligns well with the nature of the environmental disturbances, which include wind, waves, and currents---factors known to significantly affect a vessel's roll dynamics. In this regard, \approach{} provides system-specific insights that can inform adaptation logic, such as adjusting controls or switching to disturbance-aware navigation strategies. 

These capabilities demonstrate the potential of the \approach{} to be integrated into feedback loops, improving resilience and environmental awareness in SARs. Future work will explore how \approach{} can contribute to the other phases (e.g., Legitimate) of such control loops.

% aligns well with the goals of self-adaptive robotic architectures 
% by enhancing both runtime monitoring and analysis. 
% such as MAPE-K or MAPLE-K in our context. 

\paragraph{SAR applicability and generalizability}
The results show that \approach{} maintains strong performance across a range of operational conditions, including different maneuver types (zigzag, turning, random) and varying environmental disturbances (wind, waves, current) in the autonomous vessel setting, as well as indoor navigation with sensor degradation in the mobile robot setting. Notably, this was achieved without modifying the model architecture or tailoring it to each scenario. While hyperparameter tuning was performed, the resulting configurations were nearly identical across both case studies, indicating consistent behavior across applications. Despite the differences in dynamics, sensing modalities, and operating conditions, \approach{} maintained high OOD detection performance. These findings highlight its broad applicability and strong potential for generalization to other robotic systems with minimal adaptation effort.
% This suggests strong potential for applying the approach to other robotic systems with minimal adaptation effort, supporting both scalability and generalizability in real-world deployments.

\paragraph{Confidence-aware interpretation}
Our analysis revealed that the \approach{}'s confidence estimates---based on predictive variance from MC Dropout---provide meaningful insights into its behavior that can potentially inform decision-making in SARs. 
For example, in the autonomous mobile robot case, a final confidence-aware output might indicate and be interpreted as: \textit{``In the following minute (from timestep 1200 to 1260), the \approach{} is confident that the robot will exhibit abnormal behavior affecting mostly the $x$ and $y$ position states.''} 
Similarly, in the maritime vessel case, the model may highlight roll rate, roll angle, and sway speed as the most affected states under uncertain conditions. These interpretable outputs offer domain experts a clearer understanding of how abnormal behavior correlates with potential OOD sources, such as environmental disturbances, enabling faster and more informed responses.

Based on the results, in the autonomous maritime vessel case, most OOD predictions were flagged as uncertain, which helps identify situations where the model is less reliable. However, in the autonomous mobile robot, predictions indicated potential overconfidence under limited training diversity. These findings suggest that while uncertainty can support interpretability, it shall be complemented by a more robust detection indicator (i.e., reconstruction error) and possibly improved with better calibration techniques (e.g., Bayesian inference). 
In the future, we plan to explore how different dropout rates affect the confidence estimates and their alignment with actual prediction reliability.

\paragraph{Challenges and insights from analyzing real robot data}
In our preliminary experiments with the PAL Robotics case, we explored the use of real robot data. 
We collaborated with practitioners to collect data that represent both normal (IND) and OOD behaviors. 
Although collecting IND data was similar to the simulator-based setup, capturing OOD scenarios in real-world environments was challenging. 
Unlike simulation, it was difficult---and potentially unsafe---to introduce sensor or actuator noise on a physical robot, unless using faulty components, which requires strict safety precautions. 
An alternative is to create OOD scenarios through environmental disturbances for autonomous mobile robots, such as operating a robot on sandy or oily surfaces that affect robot motion, or introducing challenging objects like transparent glass obstructions. 
We noticed that these scenarios often involve multimodal data, such as combining camera images and lidar data. 
Given that our approach relies on a transformer-based architecture, extending it to support multimodal OOD detection is a promising direction for future research.

\section{Related Works}
We compare our approach to related work across four key aspects: the use of DTs for anomaly detection, proactive and confidence-aware OOD detection, embedded explainability, and conceptual advances over our prior work.

\paragraph*{Digital Twins in Self-Adaptive Robots}
Recent studies have applied DTs to support runtime monitoring, fault diagnosis, and control adaptation in self-adaptive robotic systems, including autonomous ships (\citet{hasan2024leveraging}), unmanned aerial vehicles (\citet{song2022digital}), self-driving cars (\citet{xiong2022design}), and mobile robots (\citet{betzer2024digital}). These works demonstrate the potential of DTs for enhancing situational awareness and operational safety; however, these studies target specific DT applications in isolation---e.g., monitoring or fault diagnosis/detection---whereas our approach leverages a transformer-based model for both runtime behavior monitoring and OOD detection in one proactive and interpretable DT framework. This unified approach supports not only monitoring but also analysis of SAR behavior.

\paragraph*{Digital Twins and Anomaly Detection} 
Recent advances in DT applications span various domains, including industrial wireless systems (\citet{moharam2025anomaly}), power-grid infrastructures (\citet{idrisov2025leveraging}), offshore wind turbines (\citet{stadtmann2024diagnostic}), and cyber-physical system frameworks enhanced by curriculum learning (\citet{xu2023digital}). 
% While these approaches have significantly improved real-time anomaly detection capabilities, proactive anomaly prediction combined with uncertainty quantification and integrated interpretability remains underexplored.
These approaches have significantly improved real-time anomaly detection by enabling continuous system monitoring and fault diagnosis. However, the use of DTs for proactive out-of-distribution detection---particularly in the context of self-adaptive robots---remains largely unexplored. Our work addresses this gap by introducing a DT-based approach that anticipates anomalous behavior through forecasting and reconstruction, thus extending the role of DTs beyond monitoring toward early detection and analysis of unseen or unexpected system states.

\paragraph*{Proactive and Confidence-Aware OOD Detection} 
Traditional OOD detection approaches typically operate post-hoc, flagging anomalies only after they have occurred~(\citet{hendrycks2016baseline, lee2018simple}). In contrast, proactive methodologies leverage predictive uncertainty (\citet{gal2016dropout}) to anticipate potential anomalies before they manifest. 
Our method advances this research by combining uncertainty quantification with reconstruction-based metrics, enabling proactive and confidence-aware OOD detection. In addition, we employ a Transformer-based DT, which improves the forecasting and detection capability of the system over traditional sequence models~\cite{ma2024research}. This integration allows the DT framework to detect deviations early while assessing its confidence, ultimately supporting more reliable monitoring and analysis in SAR.

\paragraph*{Explainability in Machine Learning} 
Interpretability is increasingly critical, especially for safety-critical applications, facilitating human trust and intervention (\citet{montavon2019explainable}; \citet{arrieta2020explainable}). 
% Conventional approaches such as SHAP (\citet{lundberg2017unified}) and LIME (\citet{ribeiro2016should}) offer general post-hoc explanations but rarely embed interpretability directly into anomaly detection processes. 
Widely used post-hoc explanation methods such as SHAP (\citet{lundberg2017unified}) and LIME (\citet{ribeiro2016should}) provide model-agnostic insights but are often computationally expensive and disconnected from the core detection pipeline. 
While our work does not propose a novel explainability technique in this regard, it incorporates a lightweight attribution mechanism directly within the proposed digital twin framework. This embedded mechanism identifies the most influential system states contributing to each OOD detection decision, thereby providing timely, context-specific interpretations and supporting more informed decision-making in SARs. 
% Our lightweight interpretability mechanism explicitly identifies contributing features, thus enabling rapid diagnosis and informed decision-making in SARs.

\paragraph*{Relation to Prior Work} Our earlier work~\cite{isaku2025digital} also proposed a DT-based framework for OOD detection in maritime systems. While conceptually similar, the current study significantly advances that line of research in several ways: 
\begin{inparaenum}[(i)]
    % \item it adopts a modular architecture that cleanly separates the DTM and DTC, supporting deployment flexibility and maintenance;
    \item it replaces the previous dual-model setup (RNN forecaster + autoencoder) with a single unified transformer model trained on a dual objective, simplifying integration and runtime execution;
    \item it extends applicability to two industrial case studies (autonomous maritime vessel and autonomous mobile robot), highlighting generalization across domains with diverse dynamics;
    \item it enhances interpretability by providing both confidence-aware and state attributions of OOD events, improving transparency for domain experts; and
    % \item it aligns more directly with the MAPLE-K loop by supporting both the \emph{Monitor} and \emph{Analyze} phases, whereas the earlier work focused primarily on monitoring.
    \item it aligns closely with the MAPLE-K loop by extending support from the \emph{Monitor} phase (in prior work) to also include the \emph{Analyze} phase.
\end{inparaenum}

% While these improvements broaden the scope and utility of the current approach, we emphasize that a direct empirical comparison between the two approaches in identical settings is needed to assess their relative performance and trade-offs.

\section{Conclusions and Future Work}
This paper presented \approach{}, a proactive out-of-distribution detection approach that integrates forecasting, reconstruction, and uncertainty estimation for self-adaptive robots. \approach{} demonstrated high effectiveness across two real-world case studies, specifically autonomous maritime vessels (NTNU) and autonomous mobile robots (PAL Robotics). \approach{} achieved up to 98\% AUROC, 96\% TNR@TPR95, and 95\% OOD F1-score in the autonomous vessel case, and 96\% AUROC, 94\% TNR@TPR95, and 89\% OOD F1-score in the mobile robot. The comparison results showed that \approach{} outperforms the RMSE baseline using forecast error alone, highlighting the added value of combining reconstruction and uncertainty in detecting OOD events. 

Given the results and lessons learned from this study, we identify the following directions for future work. 
\begin{inparaenum}[(i)]
    \item Extend the approach to support multimodal data fusion by including additional sensor inputs (e.g., LIDAR, camera), enabling richer representation of robotic behavior and context-aware OOD detection.
    \item Investigate uncertainty calibration techniques (e.g., Bayesian calibration~\cite{kuleshov2018accurate}) and conduct empirical evaluations on how factors such as dropout rate affect the reliability and interpretability of confidence estimates.
    \item Apply the proposed method to additional self-adaptive robots to further reinforce its generalizability and scalability across domains.
    \item Conduct a direct empirical comparison with our prior DT-based approach under identical settings---e.g., same datasets, feature sets, and evaluation metrics---to better understand their relative performance and trade-offs.
\end{inparaenum}

\section*{Acknowledgments}
% The Norwegian Ministry of Education and Research supports Erblin Isaku's PhD work reported in this paper. 
This work is supported by the RoboSAPIENS project funded by the European Commission’s Horizon Europe programme under grant agreement number 101133807.

\bibliographystyle{IEEEtranSN}
\bibliography{refs}

\end{document}